\documentclass[10pt]{article}   
\usepackage[utf8]{inputenc}
\usepackage[english]{babel}
\usepackage{graphicx}  
\usepackage{placeins} 
\usepackage[letterpaper, left=1.2in, right=1in, top=1in, bottom=1in]{geometry}
\usepackage{setspace}  
\usepackage{times}  
\usepackage[explicit]{titlesec}  
\usepackage[titles]{tocloft}  
\usepackage[square, numbers]{natbib}
\usepackage[bookmarks=true, hidelinks]{hyperref}
\usepackage[page]{appendix}  
\usepackage[normalem]{ulem}  
\usepackage{wrapfig}
\usepackage{ar}
\usepackage{siunitx}
\usepackage{url}
\usepackage{xcolor}
\usepackage{amsmath}
\usepackage{hyperref}
\usepackage{todonotes}

\usepackage[capitalise]{cleveref}
\usepackage{amssymb}
\usepackage[inline, shortlabels]{enumitem}
\usepackage{makecell}
\usepackage{booktabs}
\usepackage{multirow}

\newcommand{\mycomment}[1]{}
\usepackage{xspace}

\newcommand{\embdi}{\textsc{Embdi}\xspace}

\newcommand{\bs}[1]{\boldsymbol{#1}}

\def\E\displaystyle\mathop{\mathbb{E}}

\begin{document}
\onehalfspacing






\title{Embeddings for Tabular Data: A Survey}

\author{ Rajat Singh, Srikanta Bedathur \\ \texttt{rajat.singh@cse.iitd.ac.in},
  \texttt{srikanta@cse.iitd.ac.in}\\ 
  Indian Institute of Technology Delhi \\
  Hauz Khas, Delhi-110016, India}

\maketitle

\section{Introduction}

\textbf{\textit{Tabular data}}\footnote{Tabular Data is also known as "Structured Data" in the literature.} comprising rows (samples) with the same set of columns (attributes), is one of the most widely used data-type among various industries, including financial services, health care, research, retail, logistics, and climate science, to name a few. Tabular data is unique in several ways and comes with its own properties and challenges, as discussed in Section \ref{sec:definition} and  \ref{sec:challenges}, respectively, making it difficult to work with. Tabular data contain heterogeneous features, i.e., a table can be a mixture of different types of data: text, numerical, and categorical, to name a few. In addition, tables have intricate inter-dependencies between columns and intra-dependencies within the column.

According to the survey \cite{stackoverflow_2022}, SQL\footnote{SQL stands for Structured Query Language, helps in accessing, managing, and manipulating the Relational Databases (RDBs) where data is stored in the form of one or more tables.}, which is used to efficiently store, query, and access the tables in a database system, is one of the most popular technologies growing among developers. Hence, tables are becoming the natural way of storing data among various industries and academia. The data stored in these tables serve as an essential source of information for making various decisions. As computational power and internet connectivity increase, the data stored by these companies grow exponentially, and not only do the databases become vast and challenging to maintain and operate, but the quantity of database tasks also increases. Thus a new line of research work has been started, which applies various learning techniques to support various database tasks (subsection \ref{sec:downstream_tasks}) for such large and complex tables.

In this work, we split the quest of learning on tabular data into two phases: The Classical Learning Phase (Section \ref{sec:classical_learning}) and The Modern Machine Learning Phase (Section \ref{sec:modern_learning}). The classical learning phase consists of the models such as SVMs \cite{svm}, linear and logistic regression \cite{regression}, and tree-based methods \cite{decision-tree, randomforest, adaboost, gradient-boost, xgboost, catboost}. These models are best suited for small-size tables. However, the number of tasks these models can address is limited to classification and regression \cite{regression}. In contrast, the Modern Machine Learning Phase contains models \cite{BERT, table2vec, embdi, tabert, urlnet, atj_net, met} that use deep learning for learning latent space representation of table entities. Deep Neural Networks come with their own set of advantages and disadvantages. Some major pros of using deep learning for tabular data include flexibility, end-to-end training, building more extensive pipelines, and the ability to handle large datasets. In contrast, the significant challenges faced by deep learning models include transparency, interpretability, and heterogeneous features. We split the models into four broad categories according to how a table is visualized. Some models treat the table as an images \cite{urlnet, converting_t_to_i} and graphs \cite{atj_net} whereas others treat the table as simple text \cite{tabert, tapas, grappa} or as table only \cite{TURL, TabNet}.


\noindent
Accordingly, this survey is structured as follows:
\begin{itemize}
    \item We initiate the discussion with definitions and preliminaries of tabular data in section \ref{sec:definition}. In this section, we talked about \textit{what} is tabular data and provided definitions of the key terms used in the survey.
    
    \item Further in section \ref{sec:related_work}, we present a detailed road-map about the methods used by the practitioners to learn representation for tabular data. In subsection \ref{sec:downstream_tasks} and  \ref{sec:dataset}, we present the list of the tasks addressed in the literature and the most common datasets used by the practitioners to train and test their model.
    
\end{itemize}


\section{Definitions and Preliminaries}
\label{sec:definition}
    
    This section covers the required background for the report and provides pointers to the original works for a more detailed explanation of the methods.
    
    As a result of the proliferation of current technology and the internet's accessibility, an enormous\footnote{In the scale of Quintillion bytes per day} amount of data is created every second. Based on various types and formats, the data can be loosely divided into the three categories (Figure \ref{fig:data_types}) below:
    
    \begin{figure}[htp]
        \centering
        \includegraphics[scale=0.5]{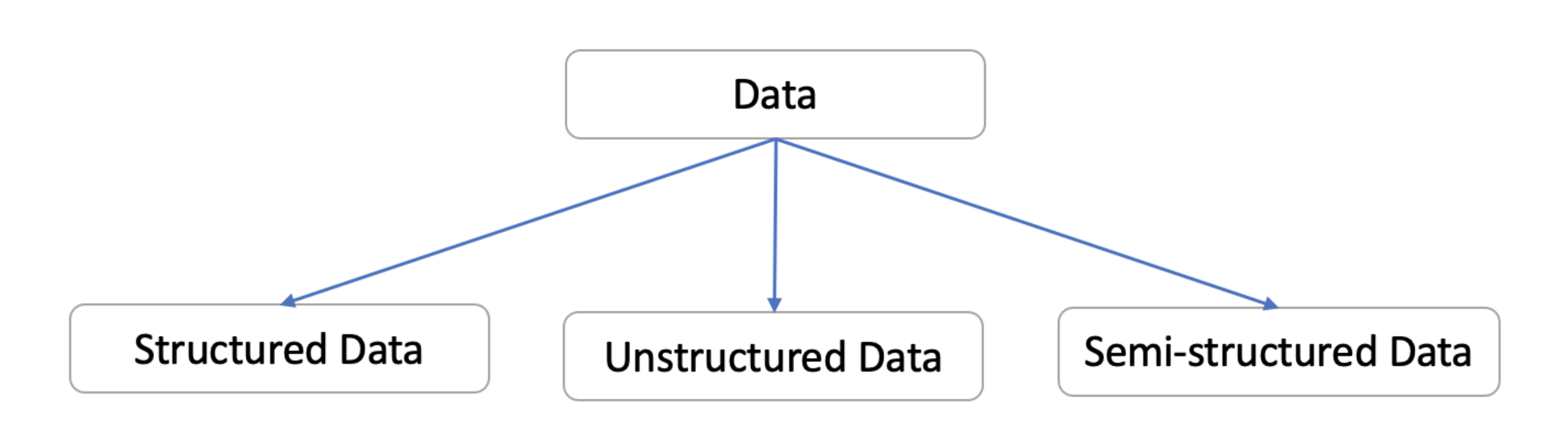}
        \caption{Classification of data.}
        \label{fig:data_types}
    \end{figure}

    \begin{itemize}
        
        \item \textbf{Structured data}: It refers to the data that has strong internal structure \mycomment{and/or is interrelated and dependent on each other, thus forming a structured space \cite{Learning_from_SD}}. Structured data are highly organized and straightforward to interpret. Structured data\footnote{Structured data is also known as \textit{quantitative data}.} can be stored in the form of tables consisting of rows and columns where data within the same column share the same semantic meaning. And each row of the table has the same syntax.\mycomment{Structured data is also known as \textit{quantitative data}.} The ideal example of structured data  is \textit{relational databases}, comprising of tables with rows and columns.
        
        \item \textbf{Unstructured data}:  It refers to the data that has no predefined internal structure and is, therefore, independent of one another. Unstructured data could be anything that can not be stored in a structured database format. Even unstructured data can have internal structures, but they are not predefined explicitly. Text files (.pdf,.doc, etc.), media (audio, video, and images) are few examples of unstructured data.
        
        \item \textbf{Semi-Structured data}: Somewhere between structured and unstructured data resides semi-structured data. They have a flexible structure, but they cannot be stored in the form of structured database format, i.e., structured tables with rows and columns. These types of data are human-crafted with markup languages. For instance, the data stored in CSV, JSON, and XML format is treated as semi-structured data.
        
    \end{itemize}

     In this report, we mainly focus on Structured data, aka tabular data, and neglect semi-structured and unstructured data as these data come with their challenges and are desirable to be discussed separately.
    
    \subsection{Tabular Data}
    
\begin{figure}[htp]
        \centering
        \includegraphics[scale=0.8]{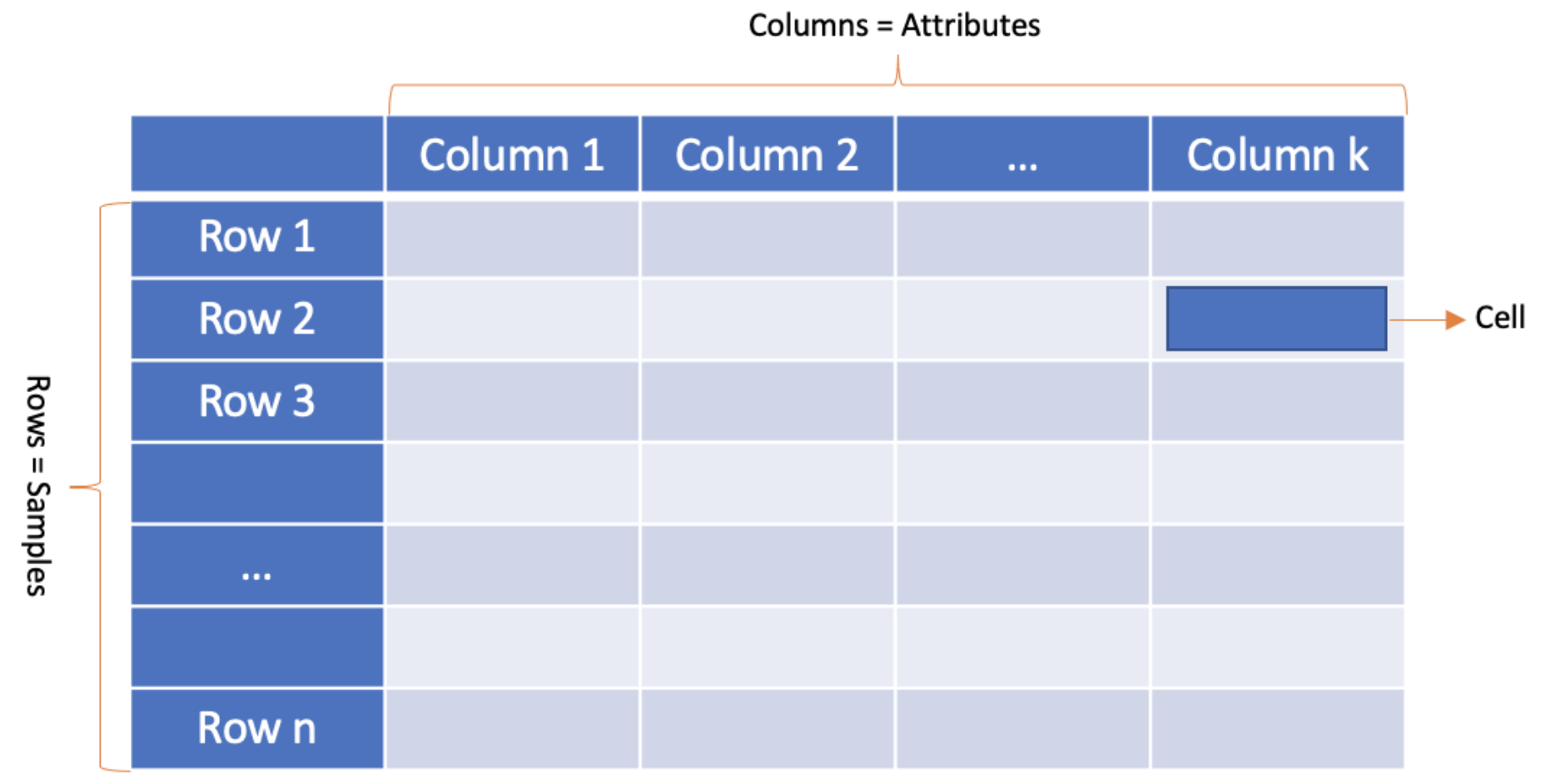}
        \caption{\textbf{Tabular data}: Representation of data in tabular form with rows and columns.}
        \label{fig:table_intro}
    \end{figure}

    In Statistics, \textit{tabular data} refers to the data that can be represented in the form of one or more tables with rows and columns (Figure \ref{fig:table_intro}). Where rows represent the samples and columns represent the attributes of those samples. Each row in the table has the same number of columns in the same order. And each column contain data of the same data type. A typical table contain table metadata, column headers\footnote{Column Headers are also termed as Column Name.}, cell information and cell metadata. A cell is a basic unit of the table formed from the intersection of a row and a column in the table.

\begin{figure}[htp] 
        \begin{center}
        \includegraphics[scale=0.6]{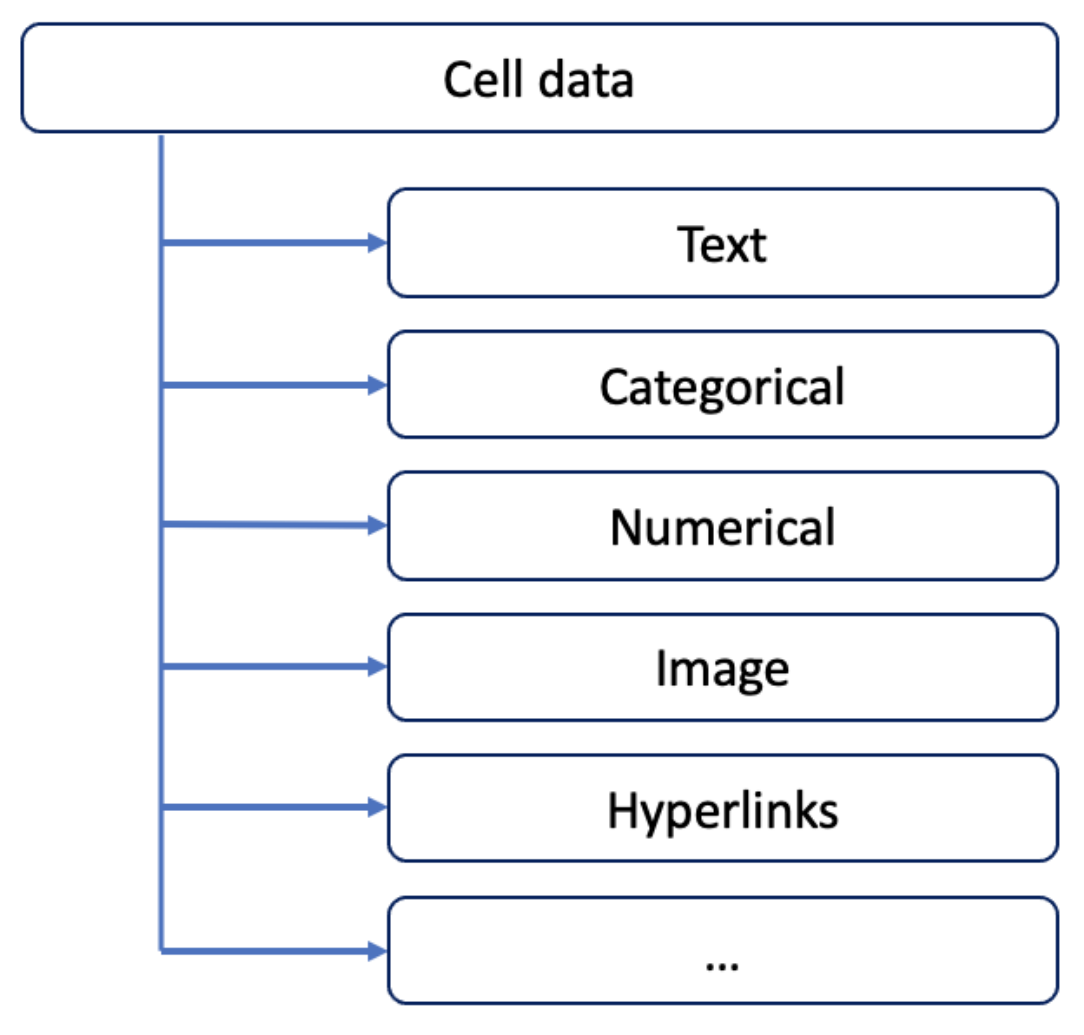}
        \end{center}
        \caption{\textbf{Cell Data}}
        \label{fig:cell_datatypes}
    \end{figure}

    Table metadata consists of a source, name, description, joins with other tables, class, and many more about the table. Table metadata provides explicit semantics about the table. At the same time, cell metadata consists of cell type, source, and so on, which is specific to each cell. A cell stores various types of data (Figure \ref{fig:cell_datatypes}), such as text data, categorical data, numerical data, image data, spatial data, hyperlinks, formulas, nested tables, and many more.

\begin{figure}[htp]
        \centering
        \includegraphics[scale=0.6]{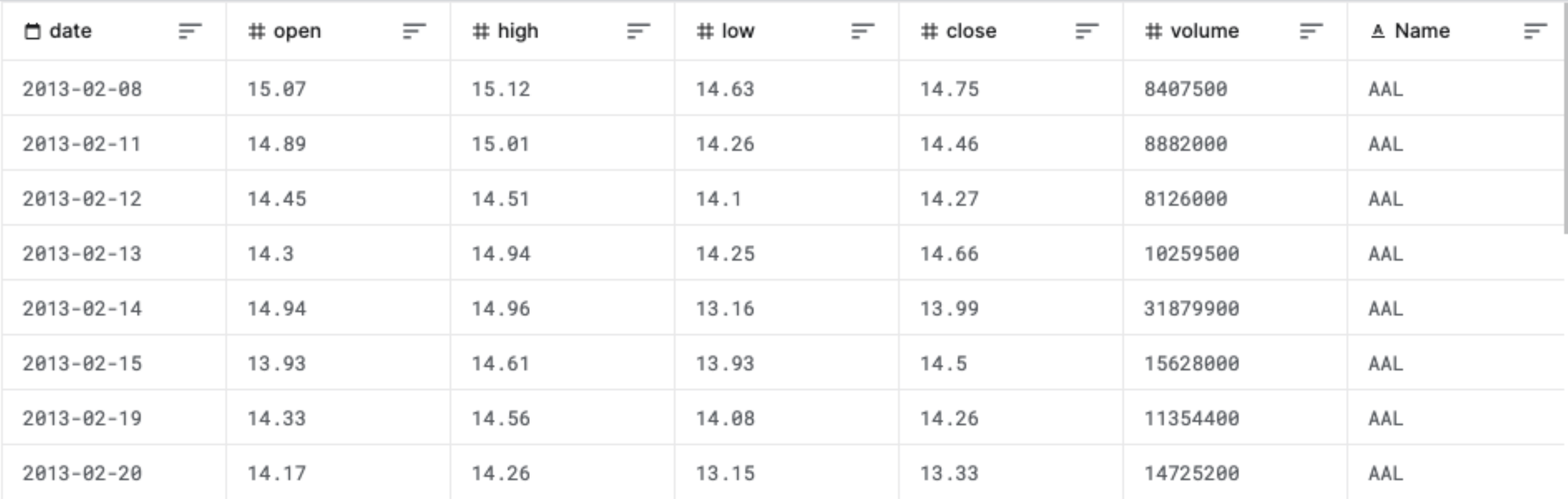}
        \caption{Snapshot of \textbf{S\&P 500 stock data} table. Major part of this table is Numerical Values.}
        \label{fig:s_p_500_stock_data}
    \end{figure}

    \begin{itemize}

        \item \textbf{Numerical data}: The majority of tabular datasets comprises of tables\footnote{Medical data, stock market, sensor data, Internet of Things(IoT) data, to name a few.} with large amounts of numerical information. Numerical data (Figure \ref{fig:s_p_500_stock_data}) provide a variety of semantic meanings, such as quantity, measurement, and ranking. It can undergo many arithmetic operations, including addition, multiplication, and proportion, to name a few.

        \item \textbf{Categorical data}: Unlike numerical data, categorical variables do not give a sense of numerical ordering. Categorical values are also called quantitative values. These values can't be measured the same as numerical values. For example, a column with \textit{gender, marital\_status, education,} etc., generally contains categorical values.

        \item \textbf{Text data}: In the tabular dataset, text data is generally a natural language text with short length and concise meaning. For example, columns with \textit{description, address, } etc., generally contain textual values. 
        
        \item \textbf{Spatial data}: Spatial data refers to information regarding the precise location of an object in an n-dimensional space. Consequently, the cells contain spatial characteristics of the item, like coordinates, dimensions, annotation, multipoint, and many more.
                
        \item Other data types, such as nested tables, hyperlinks, formulas, visual formats, images, and time can also be inserted into the table's cell. Different data types confer distinct properties on tables, which should be processed and managed accordingly.
    
    \end{itemize}

    \textbf{Properties of Tabular Data}: Tabular data has unique properties that differentiate it from other data types, such as text. Here are some fundamental characteristics of tabular data:

    \begin{itemize}
        \item The atomic entity of the table is the cell, which contains nuclear values.
        \item Each column contains values of the same type.
        \item Tables are rotation invariant; reordering the columns is inconsequential.
        \item Each row within the table is distinct, there is no meaning of repeated rows in the table.
    \end{itemize}

        
    \subsection{Embeddings}

        \textit{Embeddings}\footnote{Embeddings are also called Representation.} maps a discrete categorical variable to a vector of continuous numbers. In Natural Language Processing (NLP) context, embeddings are continuous, low-dimensional vector representations of high dimensional vectors. In general, if an embedding is learned correctly, then it can capture the semantics of the data by placing similar data points closer and dissimilar data points farther apart from each other in the latent space. Plenty of work \cite{Word2vec, BERT, word_emb, on_emb_for_NF} has already been done to find the effective embeddings for the given data. 
        

    \subsection{Attention based Encoders}
\begin{figure}[t]
            \centering
            \includegraphics[scale=0.55]{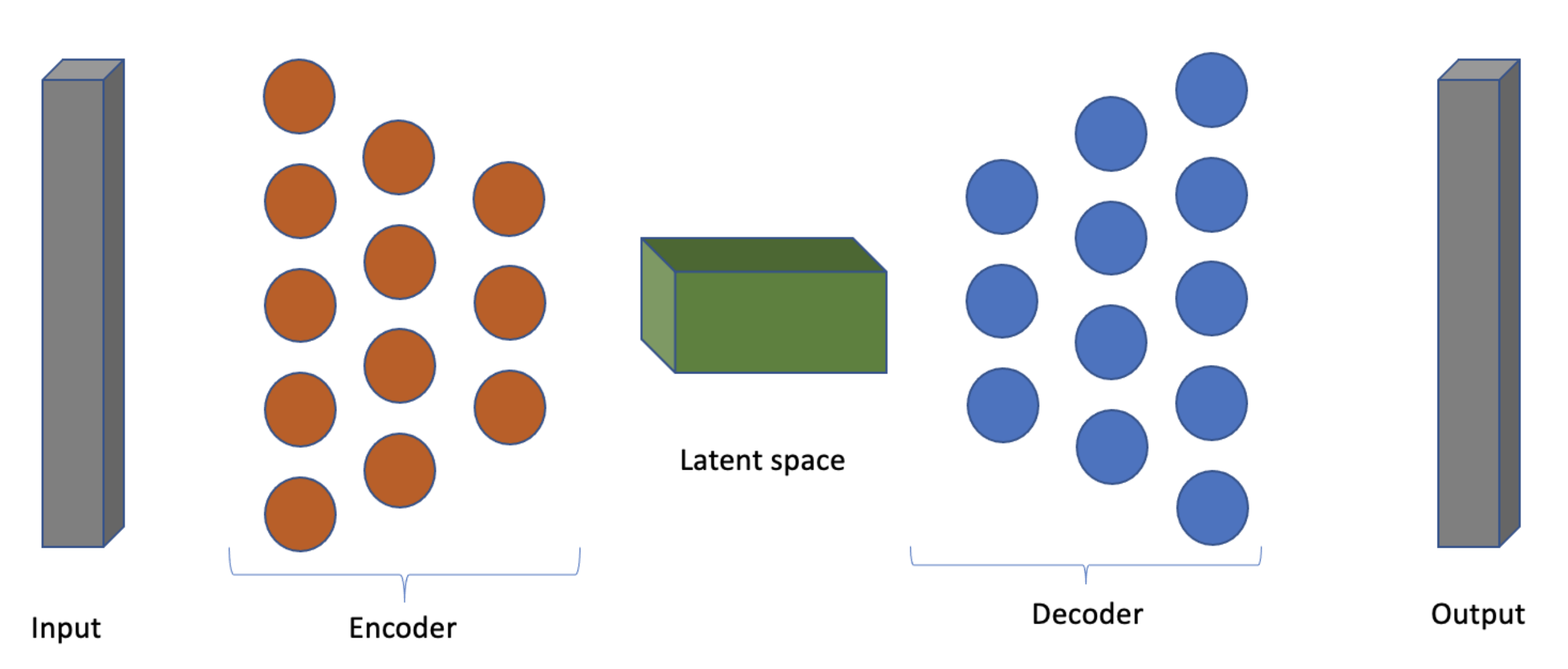}
            \caption{Typical Encoder Decoder Architecture}
            \label{fig:encoder_decoder}
        \end{figure}

        Self-attention-based \footnote{Self-attention, sometimes called intra-attention.} models \cite{transformer} have done very well at sequence-to-sequence tasks, which are used in natural language \cite{BERT, reltran, postran}, recommendation systems \cite{sasrec, bert4rec, tisasrec}, and time series \cite{thp,sahp,karishma}.  In particular, the architecture of \textit{transformer} described in \cite{transformer} involves an Encoder-Decoder (Figure \ref{fig:encoder_decoder}) architecture. An encoder is a network that takes a sequence of inputs and maps them to a latent representation with $d$ dimensions. This latent representation keeps the features that matter the most for reconstruction and discards the features that aren't needed. Further, this latent representation is passed to the Decoder to generate the input sequence. In \cite{transformer} author proposed advanced dot product attention, i.e., multi-head attention. It takes the use of query, keys, and values as input. Specifically, it uses dot-product attention defined as:
        \begin{equation}
        f_{\mathrm{Attn}} (\bs{Q}, \bs{K}, \bs{V}) = \mathrm{softmax} \left( \frac{\bs{Q}\bs{K}^{\top}}{\sqrt{D}}\right) \bs{V},
        \end{equation}
        Where $f_{\mathrm{Attn}}$ is the function to calculate attention weights, $\bs{Q}, \bs{K}$, and $\bs{V}$ represent queries, keys, and values respectively. 
        
\section{Learning on Structured Data: Challenges}
\label{sec:challenges}

\begin{itemize}
    \item[C1] \textbf{Data Quality}: It is the most common challenge faced by researchers while dealing with real-world tabular datasets. There are various reasons for low-quality of the tabular data; some of them include the imbalanced distribution of classes \cite{class-imbalance-problem}, missing values in the table \cite{Improving-DL-missing-values}, erroneous and inconsistent data present in the table \cite{anomaly-det-review}, outliers \cite{synthesizing-GAN}, and many more. The majority of the models are negatively affected by the poor quality of the data, thus opening a new domain for research to minimize the effect of poor-quality data on ML models. 
    

    \item[C2] \textbf{Complex dependencies between tokens}: Tabular data often have complex inter and intra dependencies between the tokens. Sometimes there is no relationship between the tokens of the table, and sometimes there are complex  dependencies. Therefore, models must learn these dependencies anew for each dataset. This makes it difficult for ML models to generalize on tabular data.

    \item[C3] \textbf{Different semantics for the same token}: Since there are complex dependencies between tokens of the table thus, a single token appearing in two places can have multiple semantic meanings. For e.g., in the IMDB database, token \textit{1991} under the \textit{date\_of\_birth} column of an actor is different from its presence in \textit{release\_year} column in \textit{movie} table. The same token under different columns has different types. Therefore, it is essential to capture the semantics of tokens in the database.

    \item[C4] \textbf{Pre-processing}: For tabular data, the performance of the model is heavily dependent on the pre-processing strategy used. Pre-processing consist of four major task: data cleaning, data transformation, data integration, and data reduction. Data cleaning refers to removing of incorrect, inconsistent, incomplete rows/columns from the tables. Data transformation refers to change in raw structure of the data, for example, in \cite{on_emb_for_NF}, numerical values are pre-processed and converted into a vector. Similarly, categorical values are processed and transformed into numerical ids. Data integration refers to joining of two or more tables when needed. Data reduction is the process of removing extra rows from the table and keep that many rows that makes the analysis easier and yet produce the same quality of result. In case of data reduction and data cleaning, pre-processing of data leads to information loss. 

    \item[C5] \textbf{Domain specific vocabulary}: Many of the existing works \cite{TURL, tabert} leverage the knowledge gathered by large language models (LLMs) like BERT \cite{BERT}, GPT\cite{gpt2}, T5 \cite{t5}, and so on. These approaches fine-tuned the LLMs with their tabular data. However, a domain-specific database has its dedicated vocabulary, and language models are not trained on those. For instance, MIMIC is a data warehouse of anonymized hospitalization information of patients admitted to Beth Israel Deaconess medical center. Multiple tables in the database have columns that contain special medical codes that represent different medical emergencies, diagnosis groups, and so on. In such cases, LLM-based solutions may turn out to be not helpful. 
    
    \item[C6] \textbf{Heterogeneous Data}: Most of the real-world tables are heterogeneous in nature, i.e., a  table can be a combination of columns with categorical values, numerical values, text values, and many more. Thus it is important that the model should understand the type of columns present in the table for a better understanding of the table.

\end{itemize}

\section{Related Work}
\label{sec:related_work}

\begin{figure}[htp]
    \centering
    \includegraphics[scale=0.7]{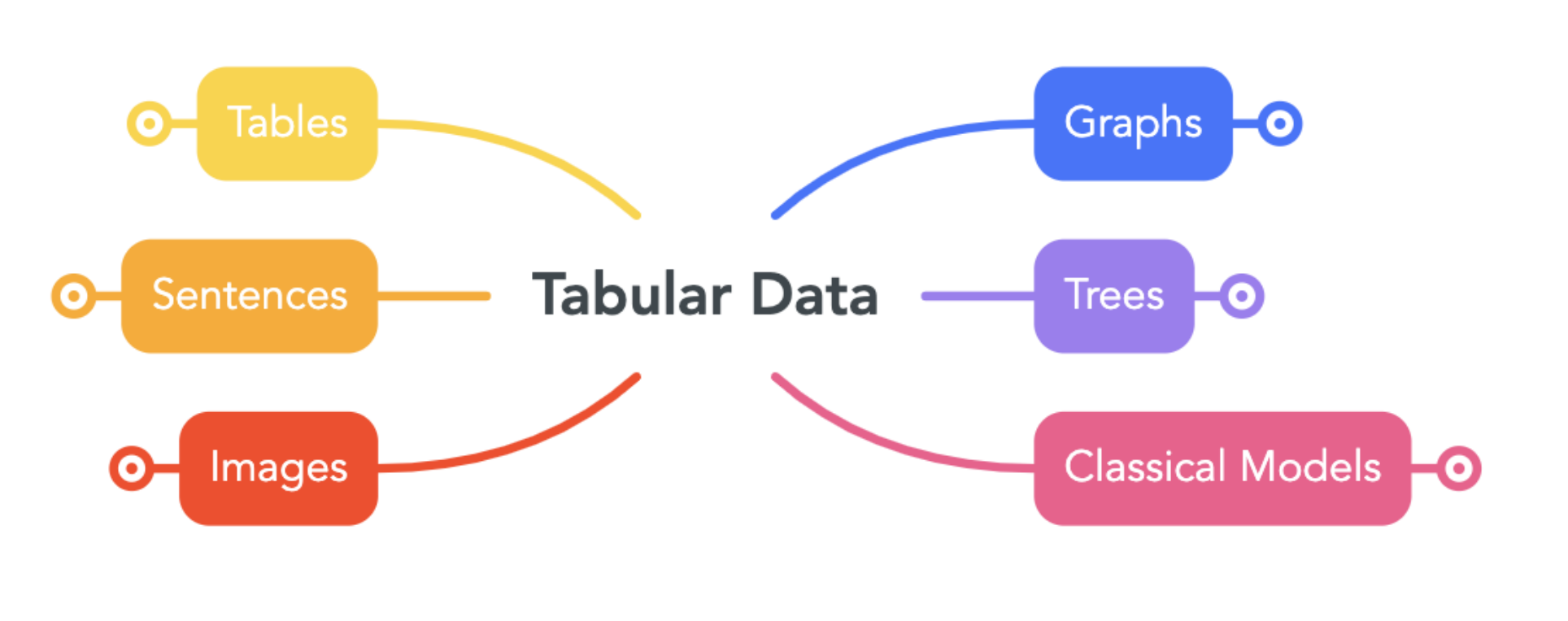}
    \caption{Explored aspects of Tabular Data.}
    \label{fig:tabular_data_classification}
\end{figure}

This section will provide an overview 
of the evolution of learning on tabular data over time. This transformation occurred in two phases; The Classical Machine Learning phase and The Modern Machine Learning phase. SVMs, Kernel Methods, Regression Techniques like Logistic and Linear Regression, Classification Techniques, and Tree-Based Methods all come under the traditional ML phase. In contrast, the Modern ML phase includes all the deep learning algorithms, such as GNNs, attention-based methods, and many more. The models that come under the classical learning phase can only capture the complex patterns in the table, thus making the model explainable. In contrast, the models used in the modern learning phase project tabular data into latent space in order to capture hidden properties and relations of the table. These models are more difficult to explain than traditional learning models.


Researchers have examined the tabular data from a variety of lenses, as illustrated in Figure \ref{fig:tabular_data_classification}. Tables \cite{TURL, TabNet}, graphs \cite{atj_net, Supervised_Learn_RD}, sentences \cite{tabert, table2vec}, images \cite{urlnet, converting_t_to_i}, and trees \cite{decision-tree, randomforest, adaboost, gradient-boost, xgboost, catboost} have all been used to represent tabular data. Each modality comes to have its own advantages and limitations. Below we tried to address all the state-of-the-art models from each modality with their limitations.



\subsection{Models}
\label{sec:models}



\subsubsection{The Classical Machine Learning phase}
\label{sec:classical_learning}

\begin{figure}[htp]
        \centering
        \includegraphics[scale=0.6]{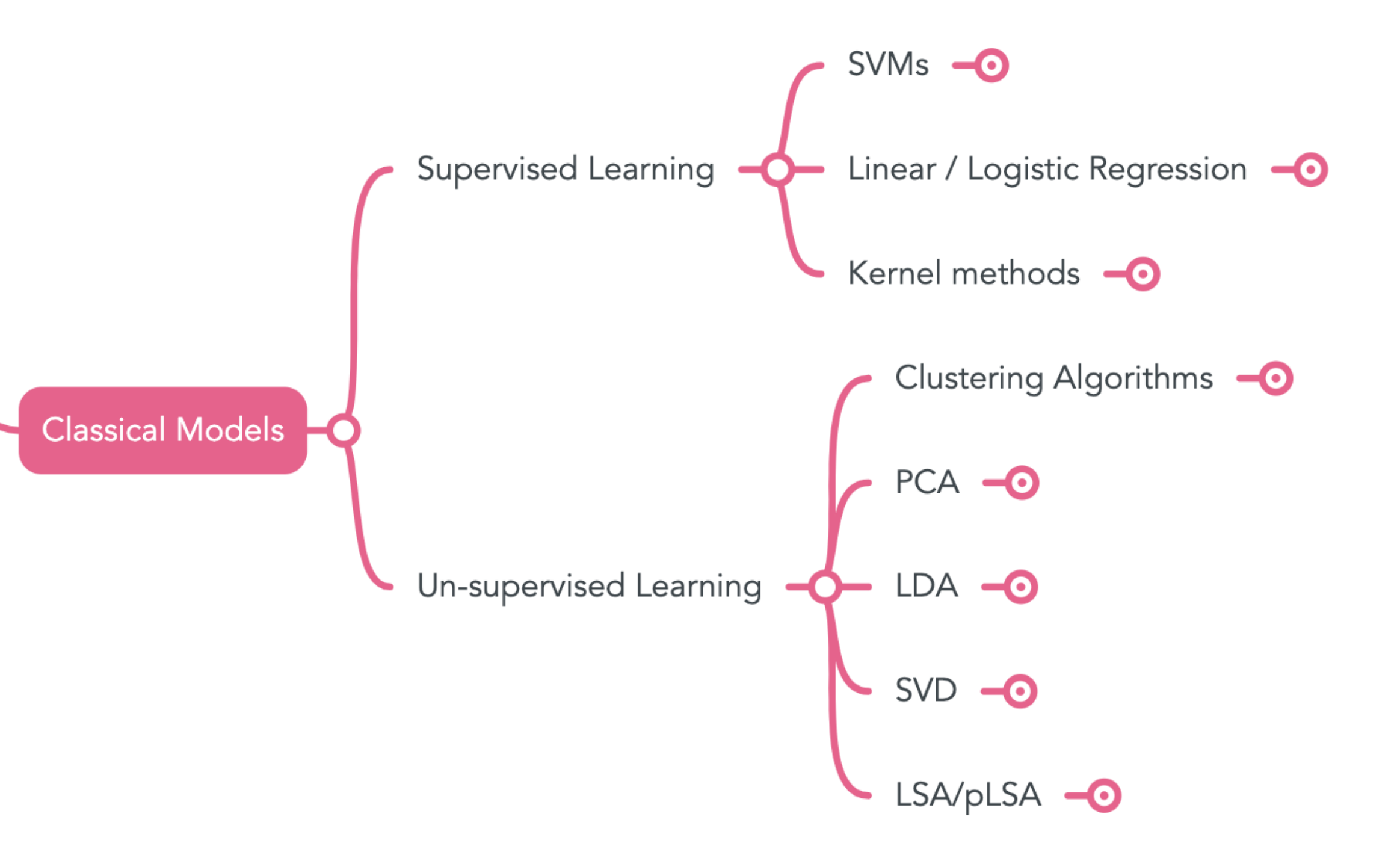}
        \caption{Classical Models}
        \label{fig:calssical_mod}
\end{figure}

\begin{itemize}


    \item \textbf{Classical Models}: Between the \textit{1950s} and \textit{1960s}, machine learning emerged as a distinct discipline. Most of the algorithms that were designed during that phase were based on probabilistic reasoning and statistics. Classical Models (Figure \ref{fig:calssical_mod}) can be broadly classified into \textbf{supervised} and \textbf{unsupervised} learning.
    Supervised Learning algorithms consist of the algorithms like SVMs \cite{svm}, Logistic and Linear Regression. In contrast, Unsupervised Learning Algorithms consist of algorithms like clustering algorithms (K-means clustering \cite{k_means}, DBSCAN \cite{dbscan}) and dimensionality reduction algorithms (Principal Component Analysis \cite{pca}, Singular Value Decomposition \cite{svd}, Latent Dirichlet allocation \cite{lda}, Latent Semantic Analysis (LSA \cite{lsa}, pLSA \cite{plsa})).
    
    \begin{figure}[htp]
        \centering
        \includegraphics[scale=0.7]{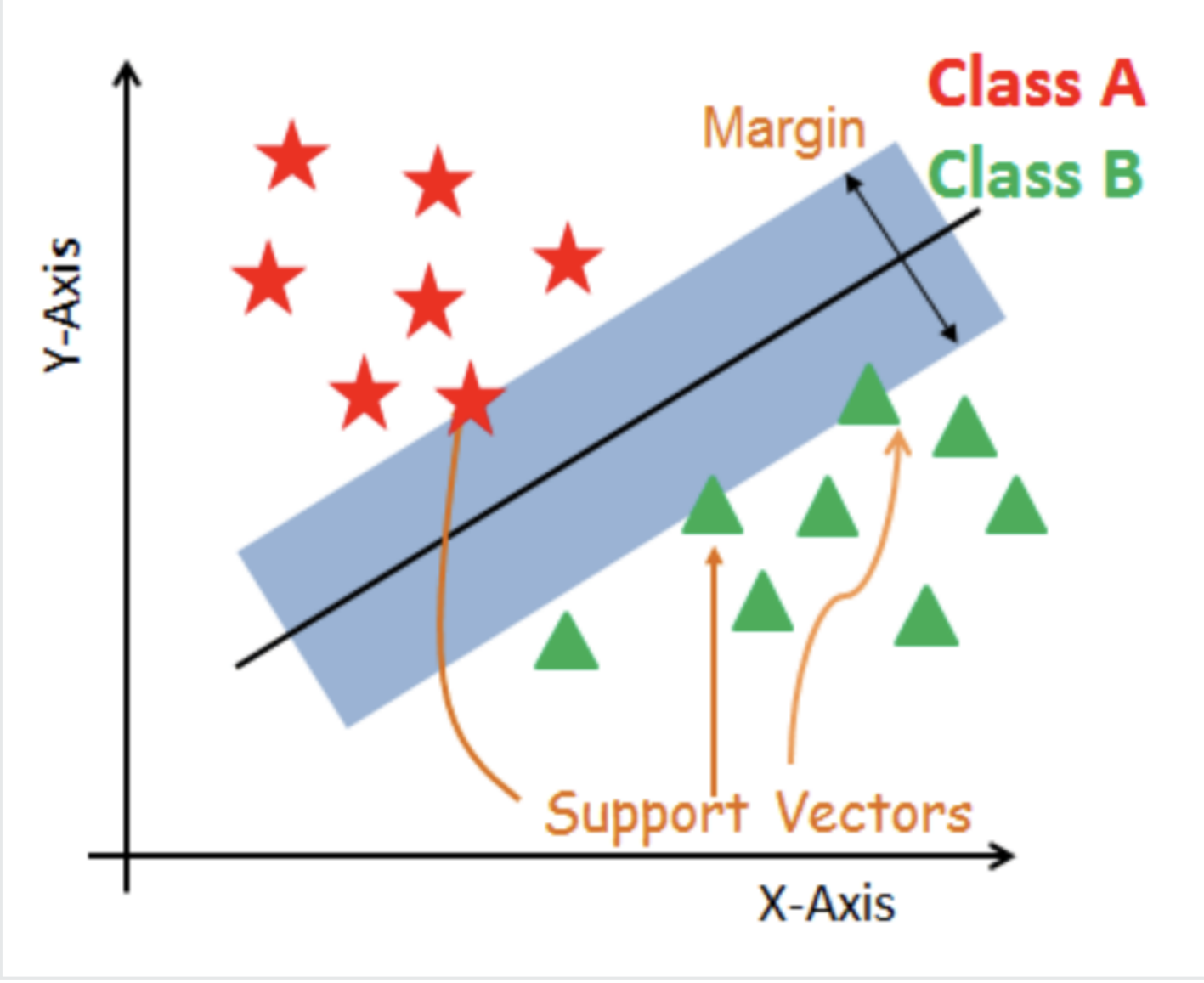}
        \caption{SVM classification (https://www.datacamp.com)}
        \label{fig:svm}
    \end{figure}
    
    Linear Regression is used for regression, i.e., predicting the continuous values, while Logistic Regression is used for classification. When there is more than one predictor variable, then it is known as multiple linear regression or multivariable linear regression.SVMs (Figure \ref{fig:svm})\footnote{Source: https://www.datacamp.com/tutorial/svm-classification-scikit-learn-python} are used for classification, regression and the detection of outliers. SVMs choose the decision boundary that maximizes the distance from all the classes' nearest data points. SVMs are broadly classified into Simple SVM and Kernel-based SVMs, where simple SVM is used in the case where data is linearly separable, and  Kernel SVMs are used when the dataset consists of non-linear data. Kernel SVMs use kernel functions\footnote{Some popular kernel functions include: Linear Kernel, Polynomial Kernel, Radial Basis Function Kernel} to address the non-linearity in the data.

    
    Clustering and dimensionality reduction algorithms come under unsupervised learning algorithms. As the name suggests, clustering tries to group data samples with the same features into one cluster, for example, k-means clustering. In contrast, dimensionality reduction algorithms try to reduce the dimensions required to represent the same data sample with minimum loss in feature. After dimension reduction, performing the downstream task over the learned representation becomes easier.
    
    
    \begin{figure}[htp]
        \centering
        \includegraphics[scale=0.5]{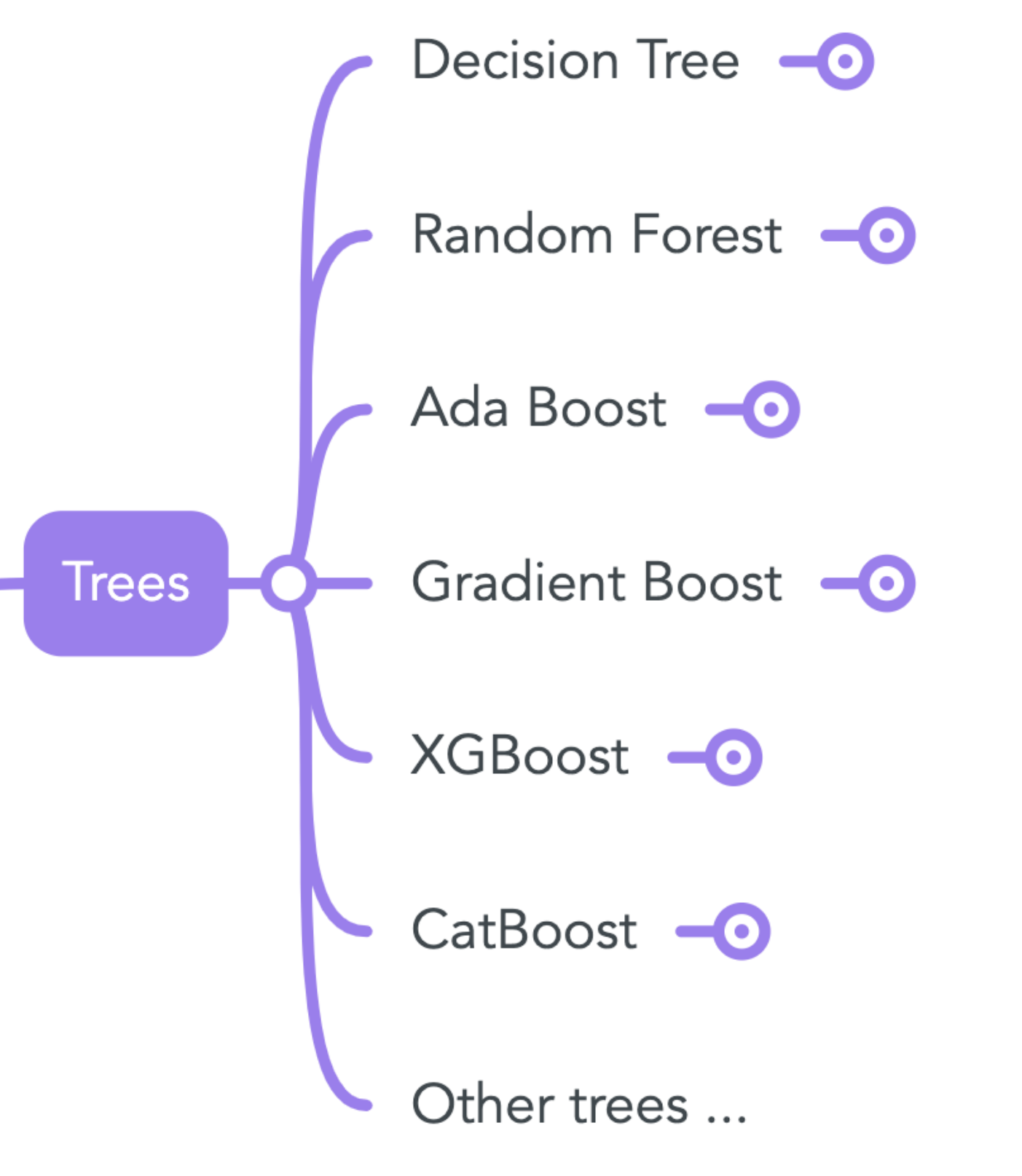}
        \caption{Tree-based Models}
        \label{fig:tree_mod}
    \end{figure}
    
    \item \textbf{Tree-based models}: Another set of algorithms that come under the classical machine learning phase is tree-based models (Decision Tree \cite{decision-tree}, XGBoost \cite{xgboost}, CatBoost \cite{catboost}, Lightgbm \cite{lightgbm}, AdaBoost \cite{adaboost}, Random Forest \cite{randomforest}). Tree-based models have been the go-to model for learning on tabular data for decades because they are not only explainable but also handle heterogeneous data.
    
    
    Tree-based models adhere to the methodology of \textbf{Decision Trees} \cite{decision-tree}, one of the first and most widely used techniques for learning discriminatory models. In general, decision trees make a statement at each step, and based on that statement; it will decide whether the statement is \textit{True} or \textit{False}. The top of the decision tree is known as Root Node, the bottommost nodes at each branch are called leaf nodes, and the nodes between the root node and the leaf nodes are called internal nodes. When a decision tree is used to classify data into categories, it is known as a classification tree. When it predicts continuous numerical values, it is known as a regression tree.

    Thus, the decision tree splits the data into two parts at each decision step. The quality of this split can be calculated using the following statistical methods: Gini impurity, Weighted Gini impurity, information gain, entropy gain, and chi-square. The most prevalent is the Gini impurity, which reduces the impurity of the decision tree. The Gini Impurity for a split is calculated as the weighted average of leaf impurity, and the leaf impurity is calculated as:
    
    \begin{equation}
        \text{Gini Impurity of a leaf (G)} = 1 - \sum_{i=1}^{n} p_{i}^{2}
    \end{equation}
    
    Where $p_{i}$ is the probability of the $i^{th}$ class in a leaf. A single decision tree does not improve performance when the dataset's structure is complex. Therefore, the assembly of two or more trees is performed to improve performance, also known as Ensemble Technique. The ensemble model is based on the idea that several weak learners can be combined to create a strong learner.

    
    Ensemble models can be broadly categorized as \textit{bagging} or \textit{boosting}. When we need to lower the variance of the decision tree classifier, we use bagging. When we need to improve the accuracy of the decision tree classifier, we use boosting. Bagging models construct different decision tree corresponding to each subset of samples chosen from the training samples with replacement. The final decision is the average of all the predictions from different decision trees, for example, Random Forest \cite{randomforest}. On the other hand, Boosting models try to improve the accuracy from prior learner to new learner by analyzing the error of the prior learners, for example, Ada Boost \cite{adaboost}, Gradient Boost \cite{gradient-boost}, XGBoost \cite{xgboost}.  
    
    \textbf{\textit{Random Forest}} \cite{randomforest} is an improvement over the decision tree. The one aspect that restricts the decision tree from being the ideal tool for learning is "\textit{inaccuracy}". Decision trees work well with the training samples but perform extremely badly with the new samples. On the other hand, random forests construct multiple random trees from bootstrapped data by randomly selecting samples from original data, which helps in visualizing multiple aspects of the dataset. The ensemble of multiple random trees is what makes random forests more effective than individual decision trees. Another advantage of a random forest over a simple decision tree is that it maintains accuracy with missing data, and it reduces the over-fitting of the model.
    
    \textbf{\textit{Ada Boost}} \cite{adaboost}, also known as Adaptive Boost, is an improvement over the random forest. The major drawbacks of random forest are that 1) it gives equal weight to all decision trees in the final answer, 2) all the decision trees are independent in the random forest, and 3) since the final prediction in the random forest is the mean of predictions from all decision trees, it is not completely accurate. In contrast, Ada Boost uses stumps, a tree with one node and two leaves, also known as weak learners. Each stump has a different weight to the final classification. Each stump is influenced by the mistake of the previous stump. Thus at each step of the Ada Boost, it creates a new learner by learning from the mistakes of the previous learner.
    
    \textbf{\textit{Gradient Boost}} \cite{gradient-boost} starts from a single leaf instead of constructing a complete decision tree or stumps. In further steps, gradient boost calculates the errors made by the previously built decision tree and builds a new decision tree on top of it based on the previous error. The key difference between the Gradient Boost and Ada Boost is that Ada Boost tries to minimize the exponential loss function that can make the model sensitive to outliers, whereas Gradient Boost can use any differential loss function. Thus Gradient Boost is more flexible and robust to outliers.  
    
    More formally, as in \cite{trees_gdbt_xgboost},  given a training dataset (Table) $\bs{T}$ = $\{x_{i},y_{i}\}_{i}^{N}$, where $x$ are features and $y$ is the target value, the goal of gradient boosting is to find an approximation function, $\hat{F(x)}$, of the actual function $F^{*}(x)$, which maps instances $x$ to their output values $y$, by minimizing the expected value of a given loss function, $L(y, F(x))$. Gradient boosting builds an additive approximation of $F^{*}(x)$ as a weighted sum of functions
     
    \begin{equation}
        F_{m}(x) = F_{m-1}(x) + \rho_{m} h_{m}(x)
    \end{equation}
    
    where $\rho_{m}$ is the weight of the $m^{th}$ function, $h_{m}(x)$. $h_{m}(x)$ are also known as weak learners.
    
    \textbf{\textit{XGBoost}} \cite{xgboost} stands for \textit{eXtreme Gradient Boost}. XGBoost is a more regularized (L1 and L2) form of Gradient Boost. It constructs the decision trees in a similar way as Gradient Boost. XGBoost is capable of parallel learning which increases its performance as compared to Gradient Boost.
    
    Researchers have done extensive experimentation over tree-based models \cite{why-tree-based-models-still-outperform-DL} and found that tree-based models are best suited for small to medium-sized tables and need a careful selection of hyper-parameters. But, nowadays majority of tabular data is of very large size, such as  WDC Web Table Corpus (233M tables), Dresden Web Tables Corpus [Eberius et al., 2015] (174M tables), WebTables [Cafarella et al., 2008] (154M tables), and WikiTables (1.6M tables). Another drawback of using tree-based models is that at each step, they perform decision-making, which destroys the semantic relation of the cell with the corresponding column and the row.  
    
    Here are some major drawbacks of using the above (the classical machine learning phase)  models:
    \begin{itemize}
        \item Since these models do not work with any type of data (like text data, categorical data, etc.) thus, a lot of feature tuning is required (such as converting categorical data to numerical data).
        \item These models cannot be used as part of the bigger pipeline.
        \item The tasks that can be performed using these models are very limited (mostly to classification and regression).
        \item Works best with small datasets as training time is very high for large datasets.
    \end{itemize}

\subsubsection{ The Modern Machine Learning phase}
\label{sec:modern_learning}
    
    The Modern Machine Learning phase is also known as "the era of Deep Learning". In this phase, deep learning algorithms outperform traditional models in numerous domains, including Natural Language Processing (NLP), computer vision, image processing, etc. By projecting the data to an intermediate representation, sometimes referred to as latent space representation ( Figure \ref{fig:encoder_decoder}), these models can discover complicated hidden features in the data. Latent space comprises a compressed representation of the data, the only feature that the decoder may use to attempt to reconstruct the input as accurately as possible. In this manner, the model can capture the hidden features in the data.
    
    Some major advantages of using deep learning models include the following:
    \begin{itemize}
        \item Since data can be projected into latent space; it is now possible to construct large and complex pipelines, which was previously impossible.
        \item End-to-end training of these pipelines is possible.
        \item In addition to classification and regression, other tasks like Question Answering (QA), missing value imputation, table-to-text conversion, and table retrieval are possible.
        \item Able to manage enormous amounts of data during training and testing.
    \end{itemize}
    
    \begin{figure}[htp]
        \centering
        \includegraphics[scale=0.66]{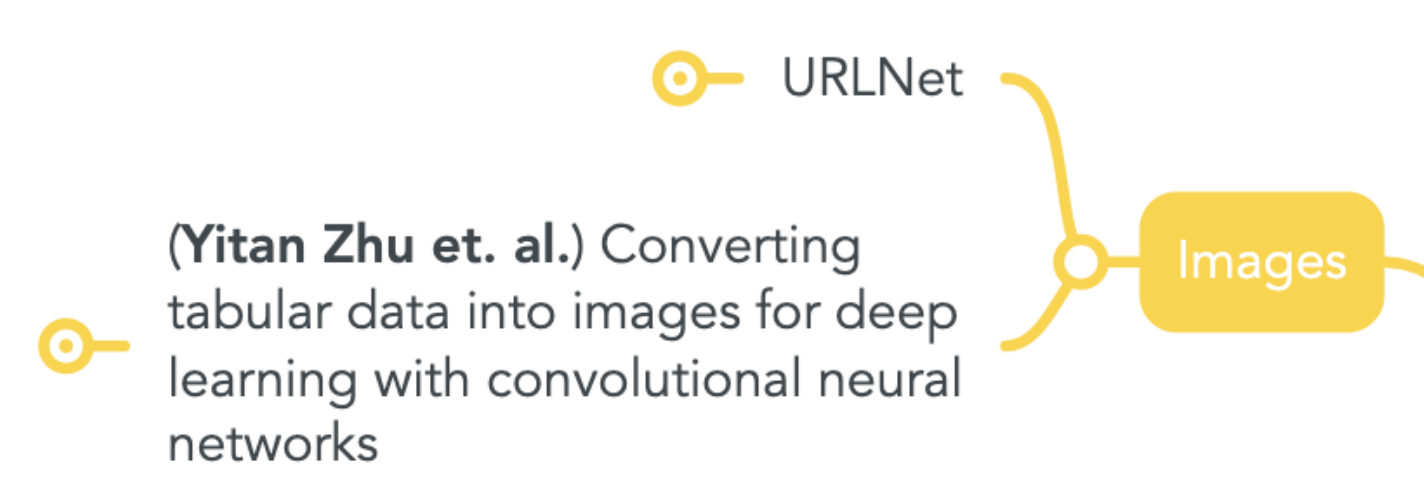}
        \caption{Image-based Models}
        \label{fig:image_models}
    \end{figure}

    \item \textbf{Image-based models}: Image-based models (Figure \ref{fig:image_models}) such as URLNet \cite{urlnet}, and Converting tabular data into images for deep learning with convolutional neural networks \cite{converting_t_to_i} take the leverage of \textit{Convolutional Neural Networks} (CNNs) for representing the tables to a latent space representation. This latent representation will be used for a variety of subsequent tasks, including classification, regression, missing value imputation, etc. Image-based models often follow the pipeline depicted in Figure \ref{fig:image_pipeline}, in which a table is transformed into an image using a certain heuristic, and that image is then passed through a CNN model to determine the latent representation of the table.
    
    \begin{figure}[htp]
        \centering
        \includegraphics[scale=0.7]{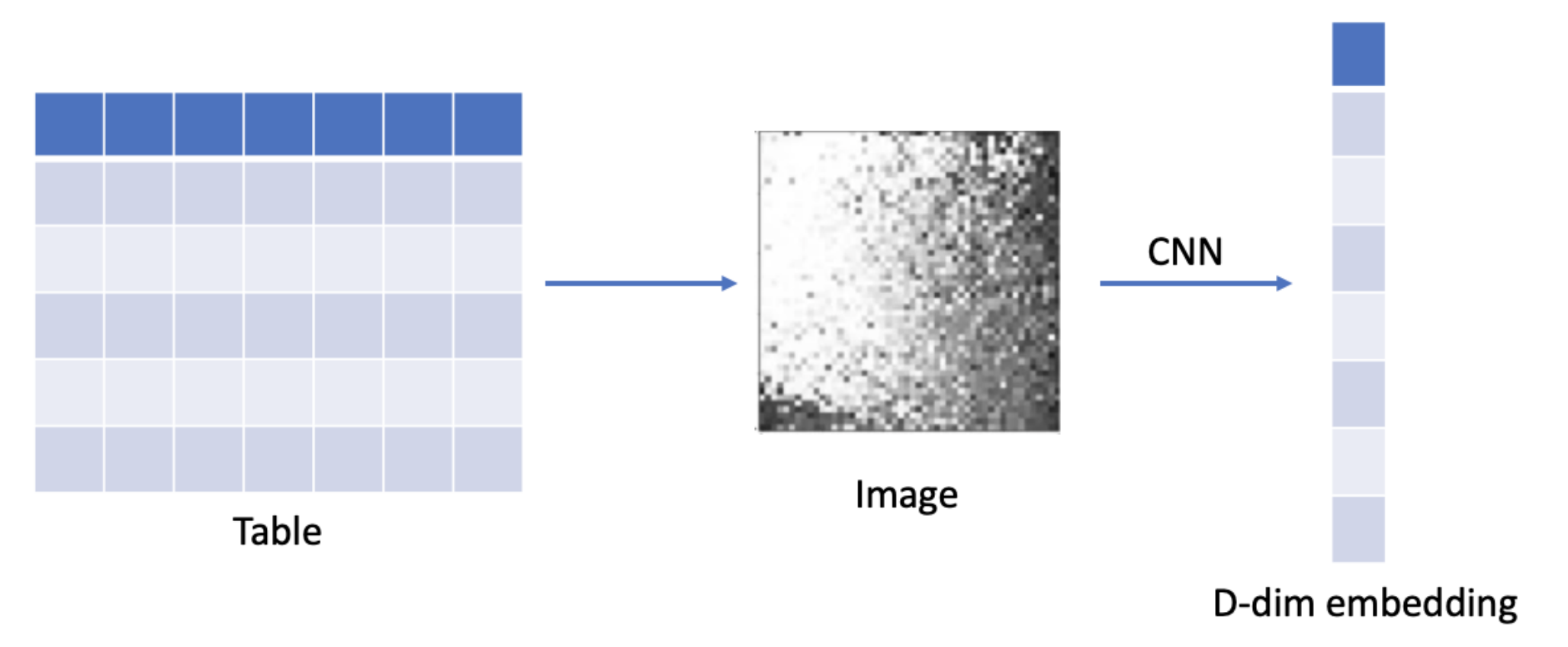}
        \caption{Image-based Pipelines}
        \label{fig:image_pipeline}
    \end{figure}
    
    URLNet \cite{urlnet} performed character level embedding to capture the spatial (sequence) level properties in the  URL and word level embedding to capture different properties of the URL such as domain, protocol, and path. Further, these word-level and character-level embeddings are concatenated to form an image-like structure. They then further pass it to the convolution layer and finally to the fully connected layer with softmax to get the probability distribution of each class. In contrast, \cite{converting_t_to_i} provides a novel approach, IGTD, for transforming tabular data to image data. 
    
    In \cite{converting_t_to_i}, authors applied the IGTD algorithm to transform gene expression profiles of cancer cell lines (CCLs) and molecular descriptors of drugs into their corresponding image representations. In addition, they trained the CNN to predict the anti-cancer treatment response on the transformed image. The authors asserted superior findings to competing models.
    
    Some major drawbacks of using image-based models for tabular data:
    \begin{itemize}
        \item A significant amount of feature engineering is required to transform the tables into images.
        \item These models cannot be generalized since they are extremely domain-specific (tables with spatial or temporal dependencies between components).
        \item Since CNNs can only capture the local relationships around a pixel; they are unable to capture the lengthy column- or row-wise relationships.
    \end{itemize}
    
    \item \textbf{Graph-based models}: In addition to images, tables can also be seen as graphs. The initial step in the pipeline for these graph-based models is to turn the table into a graph (bi-partite graph, hyper-graph, multi-graph, etc.) using some heuristics. After constructing the graph, a GNN is trained on these graphs. GNN stands for Graph Neural Networks. ATJ-Net \cite{atj_net} and \citet{Supervised_Learn_RD} use the GNN-based technique to learn tabular data embeddings. Typically, GNN-based models initially associate the graphical ideas of nodes, edges, and vertices with the table's defining attributes of rows, columns, and cell entities. They were followed by training GNN over the constructed graph to perform learning over the graph and project the graph to a latent space representation. 
    
    \begin{figure}[htp]
        \centering
        \includegraphics[scale=0.4]{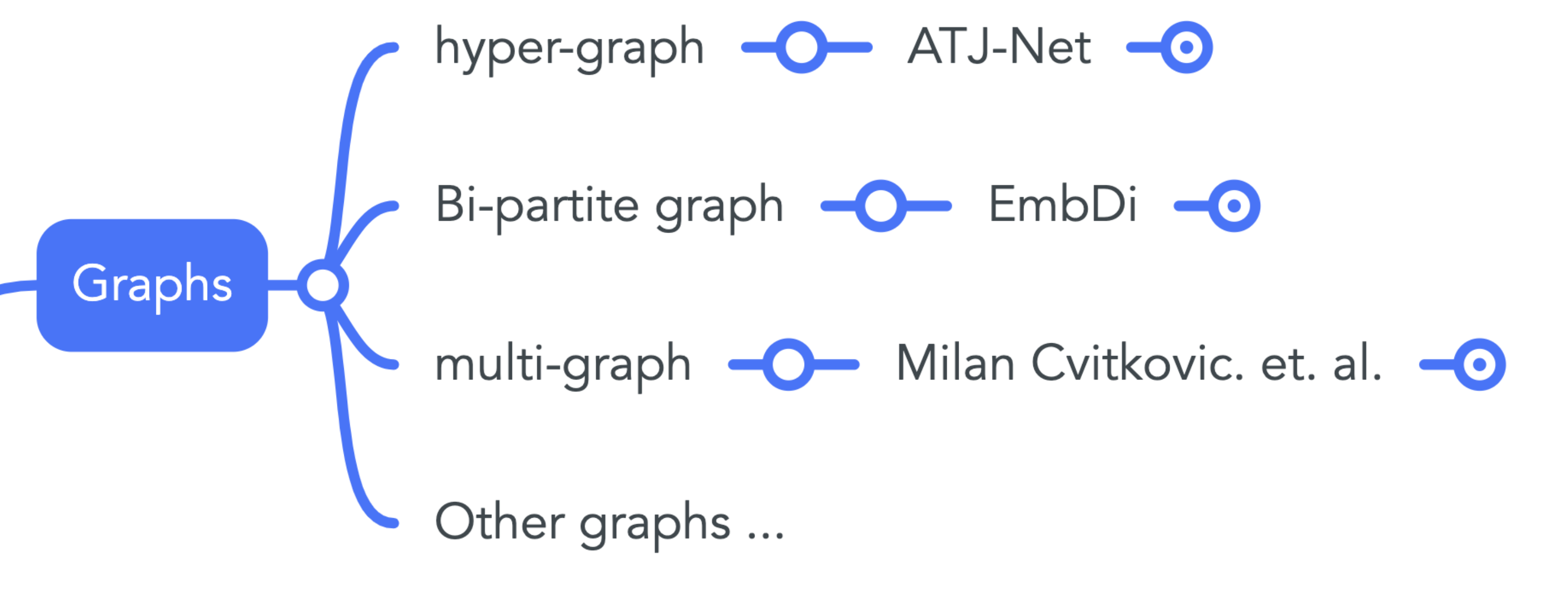}
        \caption{Graph-based Models}
        \label{fig:graph_models}
    \end{figure}
    
    The state-of-the-art model $\embdi$(\cite{embedding_of_HRD}) builds a tripartite graph with each entity connected to the $\textit{row\_ids}$ and $\textit{column\_ids}$ of the table in which it resides. Furthermore, it generates sentences using a random walk over the tripartite graph and learns the embeddings of the entities. Sentences generated by random walks on a tripartite graph may contain entities that are not directly present in the same row or column of the table. $\embdi$ learns embeddings over the generated sentence corpus by using the standard NLP-based embedding method, word2vec. Thus $\embdi$ treats the table as a graph, but at the same time, it performs a random walk to construct the sentences, which puts it under the section where we treat tables as sentences.
    
    
    ATJ-Net \cite{atj_net}, on the other hand, treats the tables as a hypergraph. As shown in figure \ref{fig:hypergraph}, hypergraph consist of vertex and hyperedge, where vertex are joinable attributes among the tables and hyperedges are tuples of the tables. In figure \ref{fig:hypergraph} joinable attribute between \textit{Review} and \textit{User} tables is \textit{"user\_id"}. Whereas, within \textit{Review} table the joinable attribute is \textit{"bus\_id"}  ATJ-Net seeks to develop a better representation of tables under multiple tables settings, even where a table is heterogeneous. This representation can be further used for many downstream tasks, including link prediction, review classification, and recommendation. To achieve this, they pre-processed the tables, and the text and image attributes in the table were converted into embeddings using transformers or ResNet. Categorical values are transferred to continuous integer values. After pre-processing, ATJ-Net turns the table into a hyper-graph with joinable attributes as vertices and tuples as hyper-edges. These hyper-graphs can be considered bipartite graphs where the tuples and the joinable attributes act as vertices to the bipartite graph. In addition, it feeds this hypergraph into a message-passing neural network (MPNN) and determines the latent space representation of hyperedges and vertices. It also employs Random Architecture Search, which automates the manual architecture design process and outperforms manually constructed models. This latent space representation is further used to perform various downstream tasks.

    \begin{figure}[htp]
        \centering
        \includegraphics[scale=0.6]{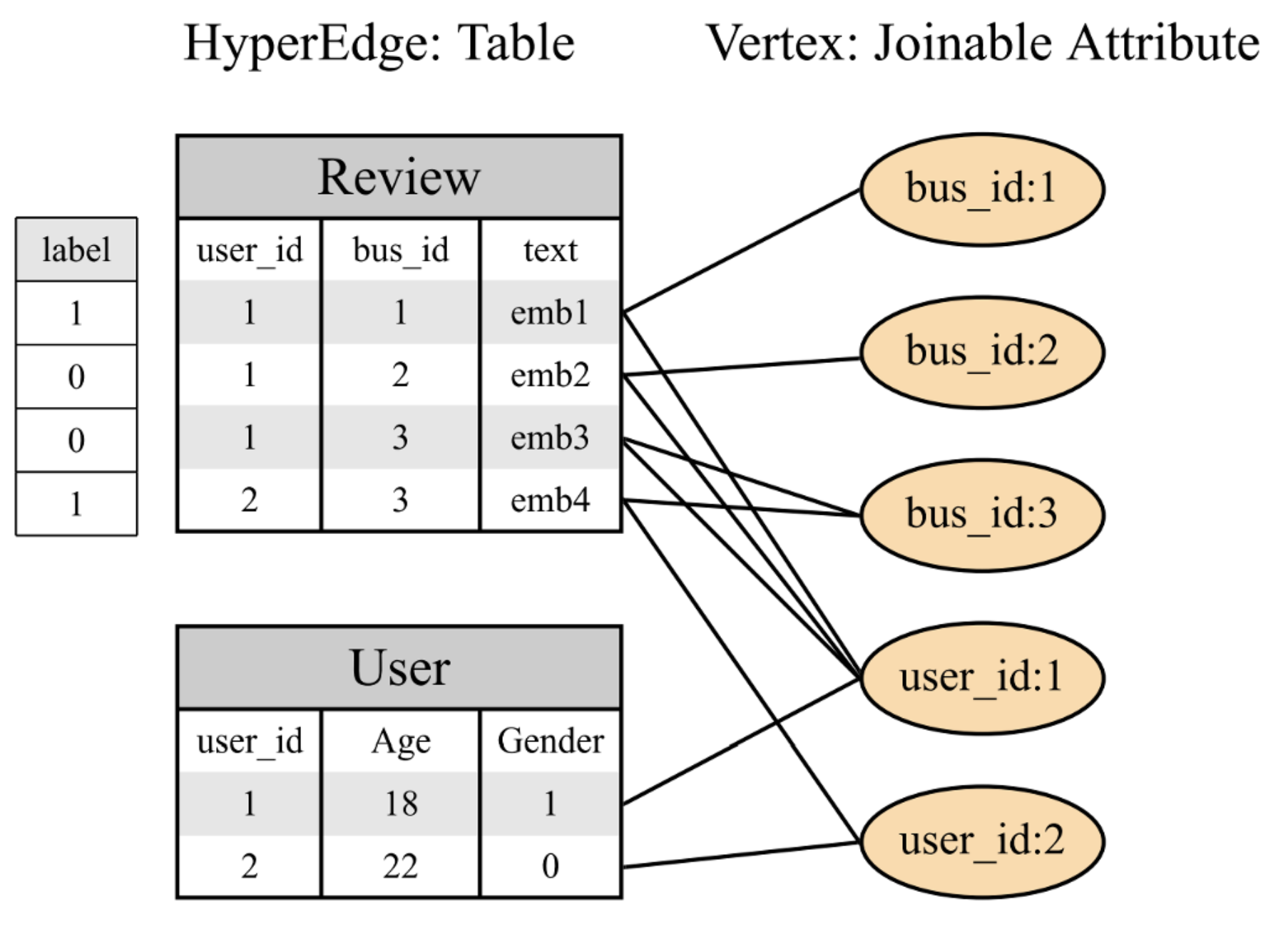}
        \caption{An example of Hypergraph \cite{atj_net}}
        \label{fig:hypergraph}
    \end{figure}

    \cite{Supervised_Learn_RD} aims to predict a single column in the table. It transforms relational tables into a directed multi-graph in which rows are viewed as nodes and foreign key references are handled as directed edges of the multi-graph. These graphs are later fed into GNN, which learns embeddings from the constructed graph. The main drawback of  \cite{Supervised_Learn_RD} is that if a table has a foreign key to itself, then it will result in selecting the entire database. In general, Graph-based models helped solve challenges that conventional NNs could not adequately address. However, extensive feature engineering is necessary for these models. Secondly, GNNs do not perform well given heterogeneous data.

    \begin{figure}[htp]
        \centering
        \includegraphics[scale=0.5]{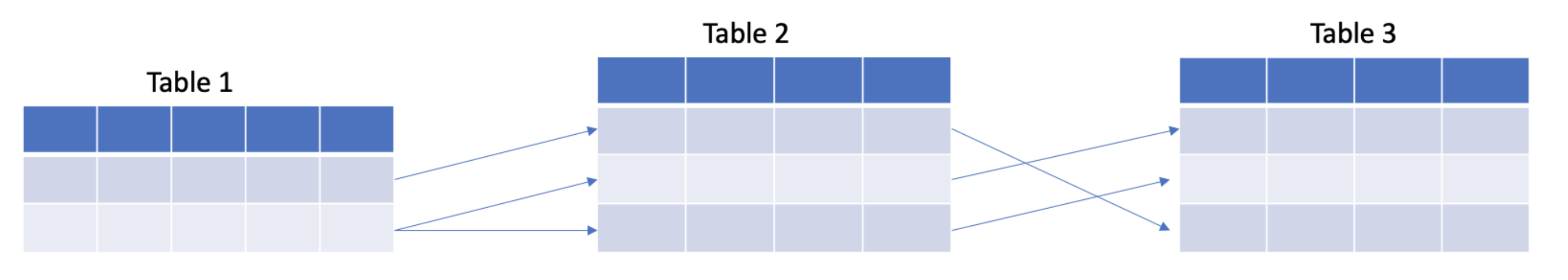}
        \caption{Multi-tabular schema}
        \label{fig:task2_1}
    \end{figure}
    
    Some major drawbacks of using graph-based models for tabular data:
    \begin{itemize}
        
        \item Tables can have different semantics for same token as shown in figure \ref{fig:task2_1} where \textit{"78"} comes in two cells with two different meaning, i.e., one represents the age \textit{"78 years"} where as another represents weight \textit{"78 Kg"}.  These tokes with different semantic meanings are hard to map in graphs.
        
        \item Extensive feature engineering is needed to represent a table into a graph.
        
        \item Since GNNs do message passing to propagate the information to the next hop node, thus there can be case when two completely independent entities affect each other's embedding.
        
    \end{itemize}
    
    

    \begin{figure}[htp]
        \centering
        \includegraphics[scale=0.8]{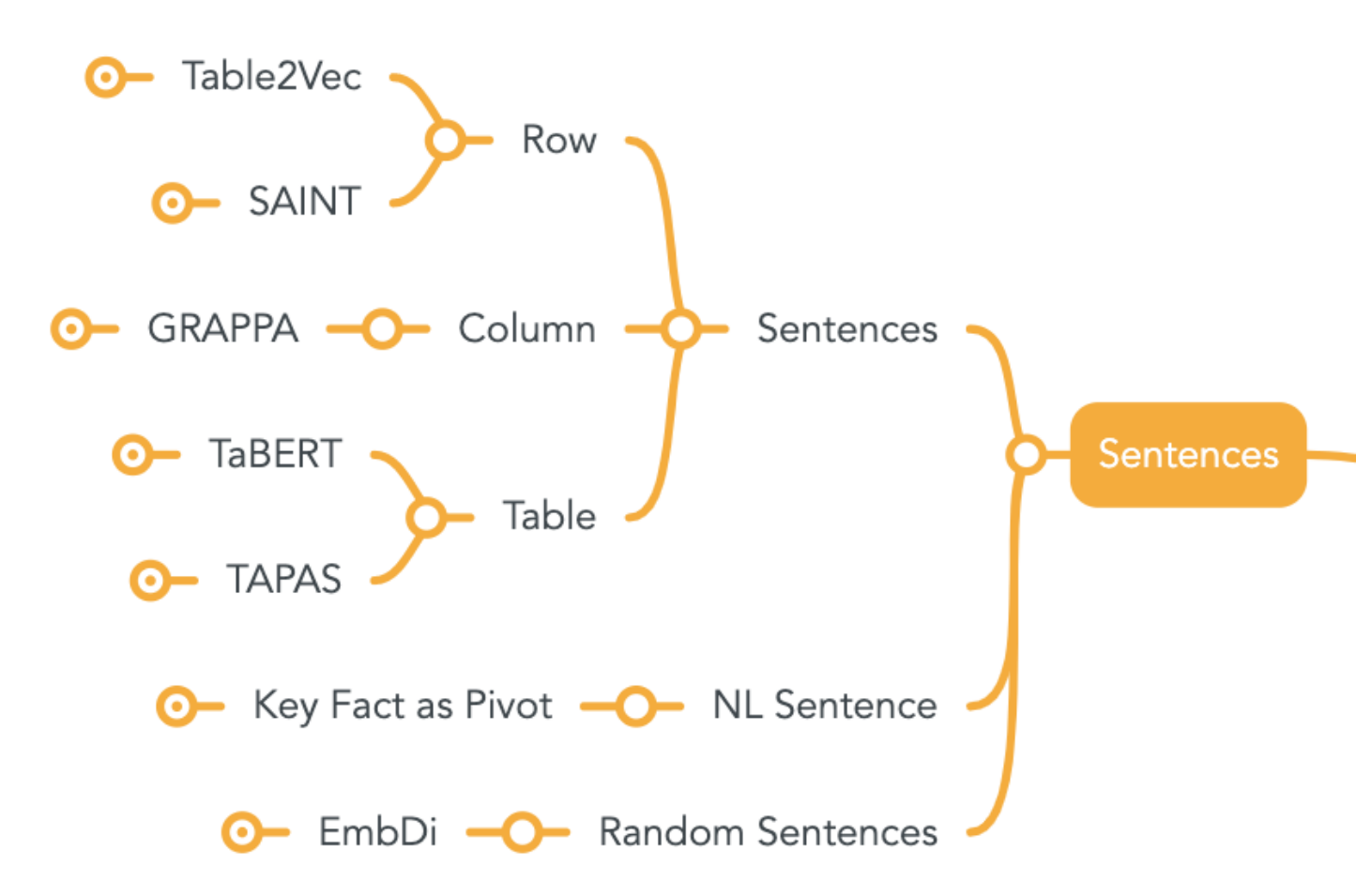}
        \caption{Table as sentence}
        \label{fig:table_as_setntece}
    \end{figure}
    \item 
    
    \textbf{Tables as collection of sentences}:
    

    With the advancement of deep learning in Natural Language Processing (NLP) tasks, many researchers have started looking at tables as collection of different types of sentences as follows:
    
    \begin{itemize}
        \item \citet{table2vec, saint} treat the row of the table as a sentence.
        
        \item \citet{grappa} treat the column name of the table as a sentence.
        
        \item \citet{tabert, tapas} treat the whole table as a sentence.
    \end{itemize}

    Table2Vec \cite{table2vec} is one of the early works that treat rows of the table as sentences. Table2Vec employs both the table data (cells) and its metadata (caption and column heading) for learning the embeddings of the entities in the tables. For learning embedding, it employs the skip-gram neural network model \textit{Word2Vec} \cite{Word2vec}. Word2Vec \cite{Word2vec}  projects the input words into a d-dimensional vector space, where similar terms are close together, and dissimilar terms are far apart. More formally, given a tuple in the table as $t_1,t_2,. . . ,t_n$, the objective is to maximize the average log probability:
    
    \begin{equation}
        \frac{1}{n}\sum_{i=1}^{n}\sum_{-c \leq j \leq c, j\neq 0 }^{}\texttt{log}p(t_{i+j}|t_i),
    \end{equation}
    
    where, $c$ is the size of training context, and the probability $p(t_{i+j}|t_{i})$ is calculated using the following softmax function:
    
    \begin{equation}
        p(t_{o}|t_i) = \frac{exp (\vec{v}_{t_{o}}^{ \intercal} \vec{v_{t_{i}}})}{\sum_{t=1}^{V} exp (\vec{v}_{t}^{ \intercal} \vec{v_{t_{i}}})},
    \end{equation}
    
    where $V$ is the size of vocabulary, and $\vec{v}_{t_{i}}$ and $\vec{v_{t_{o}}}$ to are the input and output vector representations of term $t$, respectively. After getting the embedding Table2Vec  use that embedding for task such as Row population, Column population, and table retrieval.
    
    \citet{word_emb} and \citet{ExploitingLI} focused on answering cognitive queries over relational data that involved locating entities that were similar, dissimilar, or analog. Similar to Table2Vec \cite{table2vec}, their approach involved interpreting rows of tables as natural language (NL) sentences and then training a word2vec model to embed entities in a latent space. \cite{Unlocking_NY_crime} aimed to use learned embeddings to identify nontrivial patterns from the database that helps in predicting appropriate policing matters. This way of simply interpreting table rows as NL sentences fail to capture the semantic relations between tables, which are often expressed via foreign key and primary key (FK-PK) pairs. 
    

    
    Apart from models that use Word2Vec, GloVe, fastText, and so on for learning representation, another subclass of algorithms uses transformers (attention-based models  \citet{TURL, tabert, grappa, saint, TabNet, tapas} ) for learning representation for the tables. 
    
    The core idea of these models is quite similar, i.e., the use of Vanilla Transformer with Attention modules, feed-forward network, and positional encoding. Except all these models extend the core model by modifying specific components at different levels, \cite{transformers_for_tabular_data}; (1) Input, (2) Internal, (3) Output, and (4) training level.
    
    Changes done while providing training data to a model are referred to as input level modifications. The significant changes that different models make is adding or removing various encoding.  TAPAS \cite{tapas} requires independent row encoding to be supplied to the model, whereas TABBIE \cite{tabbie} use column encoding. Other input modifications include the manner input is created; for instance, EmbDi \cite{embdi} employs random walk to construct sentences that are fed into the transformer model, whereas Table2Vec \cite{table2vec} linearizes the entire table to produce a sentence. Apart from this, the way input is being encoded, such as \cite{on_emb_for_NF} uses particular numerical encoding for numerical features, while \cite{table2vec} uses simple word2vec \cite{Word2vec} encoding, also falls under the input level.
    


    The internal level modifications are implemented to increase the model's awareness of the tabular structure. Researchers generally make modifications to the attention module to make the model structure conscious. For instance,  TaBERT \cite{tabert} uses both vertical and horizontal attention and SpanBERT \cite{spanbert} use horizontal attention.
    
    
    Changes done in the output level are related to the task that is being performed using that model. TAPAS \cite{tapas} use an additional fully connected layer over [CLS] token for predicting the Aggregation operator used in the query, whereas \cite{spanbert} uses a 2-layer feed-forward network with GeLU activations for predicting the \textit{span} of masked text.
    
    Changes done at the training level are linked with pre-training tasks and pre-training objectives. The pre-training task is used for end-to-end learning, where the model tries to reconstruct the correct input using the incorrect one. The most common pre-training task includes Mask Language Model, where the model masks random features from the input, and the task is to predict that feature; for example,  TaBERT \cite{tabert} uses column name and type masking, and SpanBERT \cite{spanbert} uses span masking. As a  Pre-training objective majority of the models try to minimize the cross-entropy loss.
    

    \begin{figure}[htp]
        \centering
        \includegraphics[scale=1]{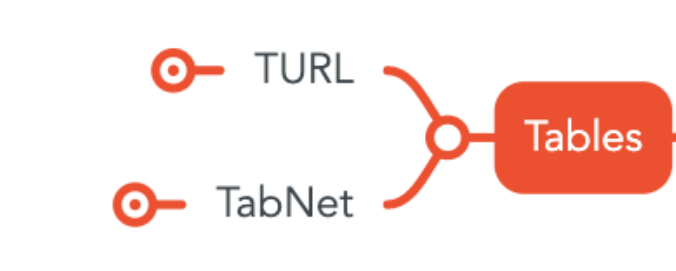}
        \caption{Table as table}
        \label{fig:table_as_table}
    \end{figure}
    \item 
    
    \textbf{Tables as table}:
    
    
    TabNet \cite{TabNet} is a transformer-based model for tabular data which does not treat a table as a sentence. TabNet tried to mimic the decision trees using attention. TabNet consists of multiple sub-networks that are processed in a sequential manner, and each sub-network act as a decision step, similar to decision trees. Each sub-network consists of two blocks: \textit{feature transformer block} and \textit{Attentive transformer block}. A few layers of Feature Transform Block are shared across all the sub-networks, and the other remaining layers are only for that particular sub-network. First, the input features are passed to the Feature Transform Block, which decides what features to pass to the next step and what features are responsible for obtaining a result at that step. The attentive transformer aggregates how much each feature has been used before the current decision step. TabNet also has a decoder that takes the encoder representation and reconstructs the feature. 
    
    The universal model TURL (\cite{TURL}), is another transformer-based framework for learning deep contextualized representations of table entities. It can be placed in both \textit{table as collection of sentences} and \textit{table as table} bucket because, it treat table as the linearize document of sentences and in the same time it uses structure-aware transformer and various types of encoding to capture the sense of table. It learns embeddings for each entity during pre-training and uses a visibility matrix to capture intra-row and intra-column relations of entities in the table. Furthermore, to get the final embedding, it used type embedding, position embedding, mention representation, and entity embedding of the entities and fed them to a structure-aware transformer with Masked Language Model (MLM) and Masked Entity Recovery (MER) as the learning objective. TURL not only uses a huge amount of metadata for pre-training but also requires an external Knowledge Base (KB) for capturing the semantics of entities in the table. It is hard to find external KB for the datasets with domain-specific vocabulary.



\end{itemize}

\subsection{Downstream Tasks}
\label{sec:downstream_tasks}

\begin{table} [t]
\centering
  \resizebox{\columnwidth}{!}{  
\begin{tabular}{l l l}
 \toprule
 Task  & Task Coverage & Representative Examples \\
 \midrule
 
 Classification & Assign class labels to problem-domain examples & \cite{saint, TabNet}\\ 
 Regression & Predicting continuous value for a sample & \cite{saint, TabNet}\\ 
 Link Prediction & Finds the relation between the given two entities & \cite{atj_net} \\
 Tables Question Answering & Retrieving table/cells from the tables for the Answer of the given NL question & \cite{tapas} \\
 Table Retrieval & Retrieving relevant table for given NL query  & \cite{table2vec}\\
 \multirow{2}{*}{Semantic Parsing}
    & Table-To-Text & \cite{tableGPT}\\
    & Table-To-SQL & \cite{grappa}\\
 \multirow{3}{*}{Table Metadata} 
    & Cell Type Detection & \cite{tabular_net}\\
    & Column Relation Detection & \cite{TURL} \\
    & Header Detection & \cite{annotating_columns_with_ptlm}\\
 Table Content Population & Populating empty cells in the table & \cite{table2vec, embdi}\\
\bottomrule
\end{tabular}}
\caption{List of Downstream Tasks for Deep Learning models.}
\label{tab:downstream}
\end{table}

In subsection \ref{sec:models}, we presented some of the works done to learn latent representation space for tabular data. However, learning latent representations alone was not our goal; rather, we wish to learn these representations so that we might perform better on a variety of downstream tasks using tabular data. This section offers a list of downstream tasks (Table \ref{tab:downstream}) that practitioners are using to evaluate the model.  
\begin{itemize}

    \item \textbf{Classification / Regression}:  Classification is a task that learns how to assign a class label to problem-domain examples. In Machine Learning, a variety of classification tasks may be encountered, and specialized modeling architectures are used to tackle each classification task. For example, figure \ref{fig:mimic_3_4} contains the schema of 4 tables from  MIMIC III dataset, and we want to predict the mortality risk of a patient given the \textit{patient details, admissions details, ICU stay and prescriptions}, is a classification task given mortality rate can be only \textit{high, low, or moderate}. Some classification tasks include Binary Classification \cite{binary_classification}, Multi-Class Classification \cite{multi-class_calssification}, and Multi-Label Classification \cite{multi-label_classification}. In contrast, regression is similar to classification, except in classification, we tend to predict the values which are continuous in nature. For example, in figure \ref{fig:mimic_3_4}, we want to predict the number of hours a patient is likely to stay in ICU given the \textit{patient details, admissions details, ICU stay and prescriptions}, is a regression task, where hours can be any real number within some range.
    
    \begin{figure}[t]
        \centering
        \includegraphics[scale=0.45]{ 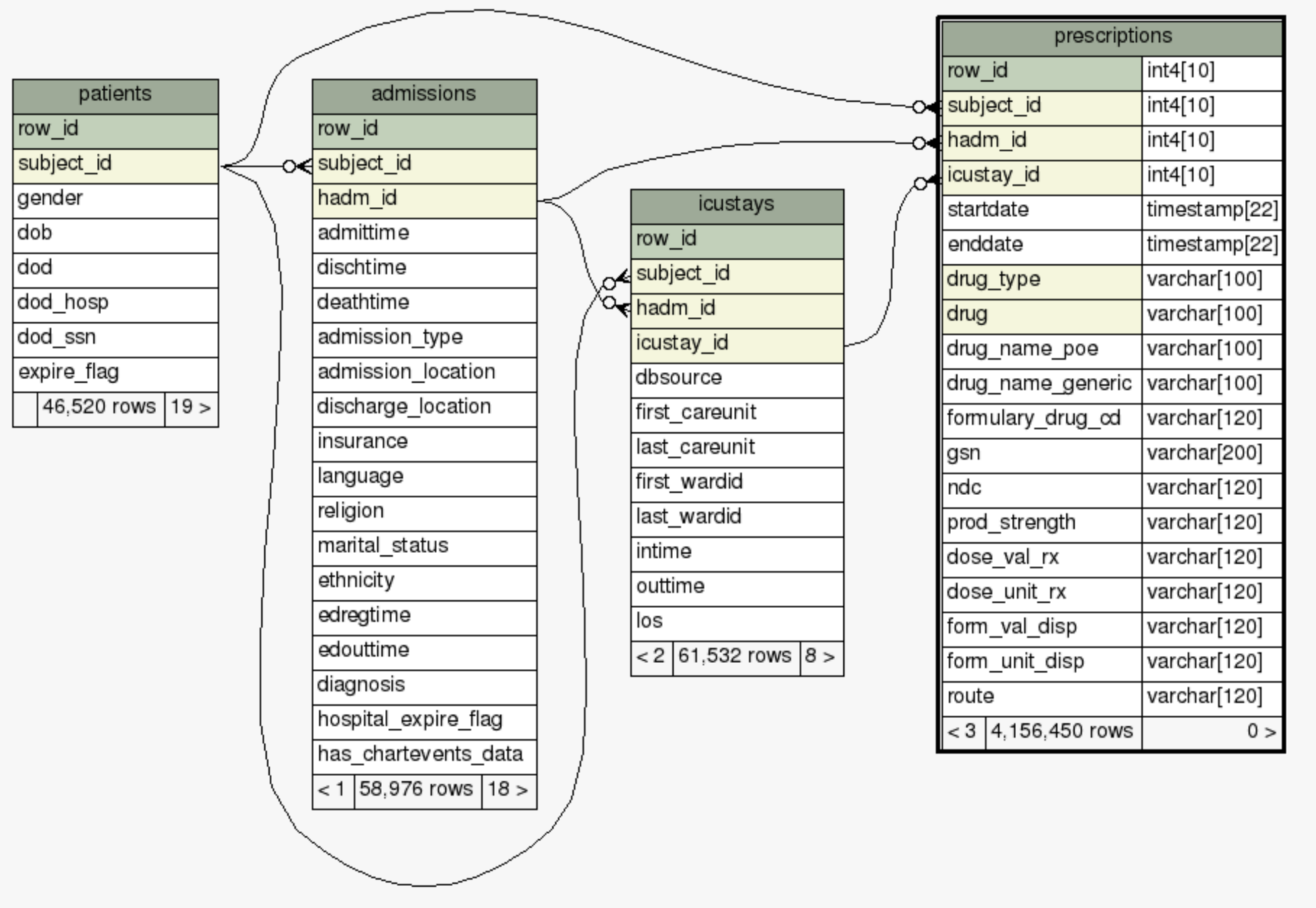}
        \caption{MIMIC III dataset schema of 4 tables. (https://mit-lcp.github.io/mimic-schema-spy) }
        \label{fig:mimic_3_4}
    \end{figure}

    \item \textbf{Link Prediction}: In recent years, social network analysis has gained considerable interest. The most important research direction in this field is link prediction. Link prediction is the likelihood of a relationship between two entities in a table. In ATJ-Net \cite{atj_net}, authors use the Aminer dataset, which is an academic, social network dataset, to train their model ARM-Net to predict the link between author and paper (as a citation) and author and author (as collaboration). In context of tabular data link prediction is equivalent to find a join or entity linking task between two tables.
    

    
    \item \textbf{Table Question Answering}: Given an input as Natural Language (NL) query, the table question answering objective is to retrieve the table/cell which contains the answer to the given NL query. For example, given the table in figure \ref{fig:wrestling} and the NL query \textit{"Which world champions had only on reigns ?"} then the model should give the result as \textit{"Dory Funk Jr."} and \textit{"Gene Kiniski"}. There are two levels of complexity in Table Question Answering, i.e., simple question answering and complex question answering. Simple question answering handles the simple lookup queries \cite{tableQA}. On the other hand, complex question answering task involves numerical values and aggregation operations \cite{tapas}. 

    \begin{figure}[htp]
        \centering
        \includegraphics[scale=0.7]{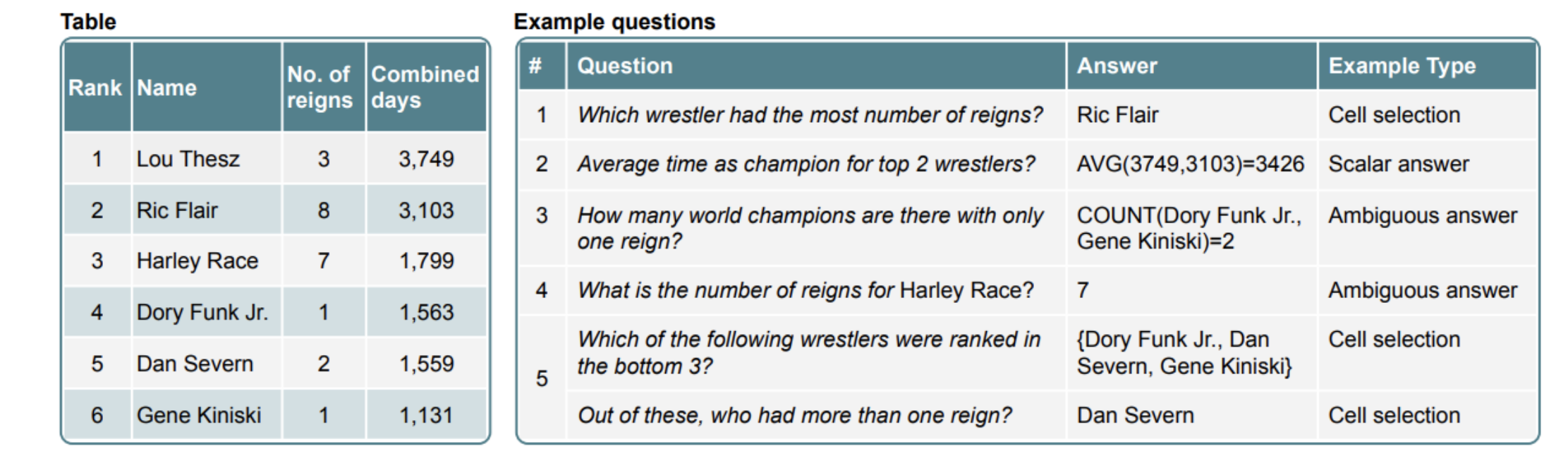}
        \caption{A table (left) with corresponding example questions (right). \cite{tapas}}
        \label{fig:wrestling}
    \end{figure}

    \item \textbf{Table Retrieval}: Table retrieval refers to the task of retrieving relevant ranked list of tables $T_1, T_2, ..., T_n$ from a collection of tables $T$, given a query utterance (figure \ref{fig:table_retrieval}). This is one of the least explored aspect of tabular data but have a significant impact. For instance, table retrieval can be used to restrict the search space for Table Question Answering. \cite{retrieving-complex-tables, open-domain-ques-answering}.

    
    \begin{figure}[htp]
        \centering
        \includegraphics[scale=0.5]{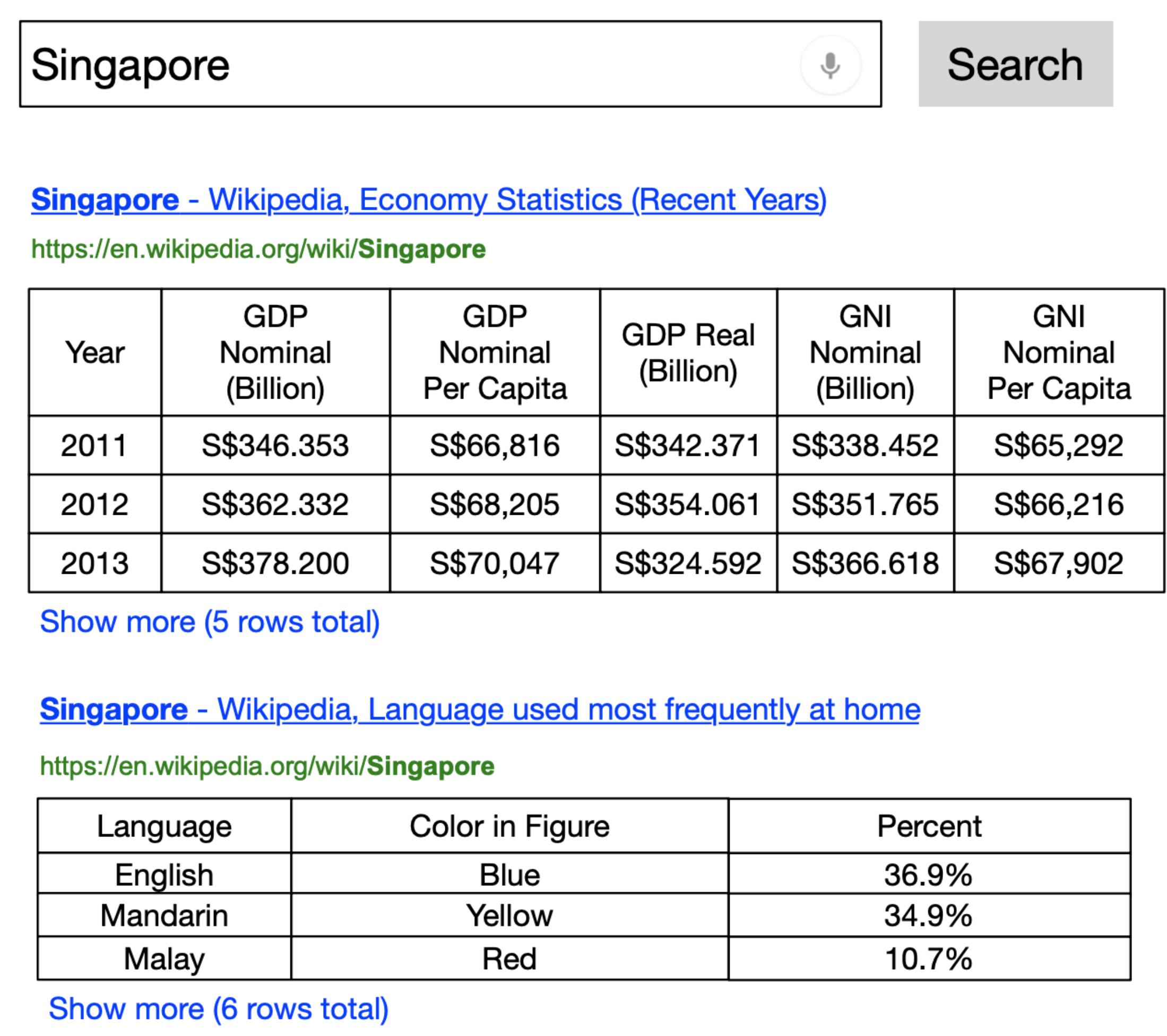}
        \caption{: Ad hoc table retrieval: given a keyword query, the system returns a ranked list of tables \cite{ad_hoc_table_retrieval}}
        \label{fig:table_retrieval}
    \end{figure}

    \item \textbf{Semantic Parsing}: Semantic Parsing translates natural language utterances into a meaningful representation. Some of the common semantic parsing tasks in tabular data include Table-To-Text conversion and Table-To-SQL parsing. Table-to-text generates textual description for the given table \cite{tableGPT}. On the other hand, given a table and the Natural Language Question as input, Table-to-SQL \cite{grappa} generate a SQL query that will retrieve the answer of the given NL query over the table.
    
    \item \textbf{Table Metadata}: Table metadata task generally relates mainly to the following tasks: \textit{Cell type, Column type, relation, and header detection}. TURL \cite{TURL} captures the relation between the entities using a visibility matrix. Whereas TaBERT \cite{tabert} encapsulates the column type by concatenating the cell value with the column type. 
    
    
    
    
    
    \item \textbf{Formula Prediction}: Given the table and the target column of the table, the Formula prediction objective is to predict the spreadsheet formula such as addition, subtraction, maximum or minimum, for the target column of the table.
    
    \item \textbf{Table content population/ Table augmentation}: Unlike table metadata where table metadata is noisy and/or missing, Table content population deals with corrupted or missing cell content. Given an input table with missing values, the objective is to predict the missing cell values. \cite{table2vec, embdi}
    
\end{itemize}



\subsection{Dataset}
\label{sec:dataset}

    In this section of the survey, we will examine numerous facets of tabular data used by practitioners. The datasets used to train and test a model plays a crucial role in comprehending the model's behavior and gaining a complete holistic view of the model. If a model is trained on an unbalanced dataset, it will be skewed toward the majority class and will not produce accurate predictions for minority-class test cases. Similarly, there are ML models \cite{on_emb_for_NF} that explicitly look into the numerical data and may not fit for categorical or another type of structured data.
    
    
    Tabular datasets may contain a single table to store the entire dataset, or they may contain multiple interconnected tables where the data is distributed across multiple tables. Based on the size of Structured Data, datasets can be classified into three categories; small, medium, and large size datasets. Small-size datasets are those datasets that have too few features ($\sim$ 5) and too few  ($\sim$ few hundreds) samples. Small-size datasets are very sparse in features and samples; thus, it becomes hard to learn any meaningful pattern from them. On the other hand, medium-size datasets contain a few hundred to a few thousand ($\sim$ 10K) samples. Extensive experimentation shows that tree-based models \cite{why-tree-based-models-still-outperform-DL} remain state-of-the-art on the medium-sized structured dataset. Finally, large-size datasets are those datasets that are larger than medium-sized datasets. Generally, Deep Learning based models are best suited for these large-size datasets. 
    
    Based on the type of entity in the tables, tabular datasets can be classified into two categories: Homogeneous tabular data and heterogeneous tabular data. Homogeneous tabular data contain a single data modality, i.e., if the table contains numerical data (\textit{Arcene dataset} \cite{Arcene}), then all the rows and columns will be only numerical. Unlike homogeneous tabular data, heterogeneous tabular data is a mixture of different modalities such as \textit{Arrthythmia} containing categorical, Integer, and real number modalities, \textit{MNIST} containing categorical and image modalities.
    
    Table \ref{tab:datasets} contains some of the popular tabular datasets and the task that is being performed using these datasets. Such as Wikipedia Tables are one of the most commonly used tabular datasets extracted by crawling over Wikipedia pages. This dataset contains around ~1.6M tables with additional information about its surrounding text, such as page caption, title, and description. Similarly, SPIDER is a famous semantic parsing dataset generally used for the text-to-SQL task. It contains tables, a natural language question (10,181), and the SQL query (5,693) associated with that NL question.

\begin{table} [t]
\centering
  \resizebox{\columnwidth}{!}{  
\begin{tabular}{l l l l}
 \toprule
 Dataset  & Task Coverage & Modalities \\
 \midrule
 Wikipedia Tables & TQA\footnote{TQA: Table Question Answering}, SP\footnote{SP: Semantic Parsing}, TCP\footnote{Table content population}, TM\footnote{TM: Table Metadata} & It contain tables and its surrounding text \\
 WDC Web Tables Corpus & TQA, SP & It contain table and its metadata and the surrounding text about the table\\
 SPIDER & SP & It contains table + NL Question + sql for the query\\
 WikiTQ & TQA, SP & It contain  semi-structured tables and question-answer pair\\
 MIMIC & TCP, Classification &  It contain large tables with domain specific vocabulary\\
WikiSQL & SP, TR\footnote{Table Retrieval} & It contain  semi-structured tables and question-SQL pair\\
Forest cover-type & Binary Classification & Table contain numerical and categorical data \\

\bottomrule
\end{tabular}}
\caption{Popular Tabular Datasets}
\label{tab:datasets}
\end{table}

    
    
    
    
    

    
        





\newpage
\bibliographystyle{unsrtnat}
\bibliography{reference.bib}


\end{document}